\pdfoutput=1

\documentclass[11pt]{article}

\usepackage{acl}

\usepackage{times}
\usepackage{latexsym}

\usepackage[T1]{fontenc}

\usepackage[utf8]{inputenc}

\usepackage{microtype}

\usepackage{hyperref}
\usepackage{url}

\usepackage{subcaption}
\usepackage{multirow}
\usepackage[shortlabels]{enumitem}
\usepackage{microtype}
\usepackage{graphicx}
\usepackage{booktabs}
\usepackage{hyperref}
\usepackage{ifthen}
\usepackage{cuted}
\usepackage{makecell}
\usepackage{tabularx}
\usepackage{multicol}
\usepackage{wrapfig}

\usepackage{amsmath}
\usepackage{amssymb}
\usepackage{amsthm}
\usepackage{dsfont}
\usepackage{multirow}
\usepackage[shortlabels]{enumitem}
\usepackage{setspace}

\usepackage{bbm}
\usepackage{hyperref}

\usepackage{amsmath,amsfonts,bm}

\def\eqref#1{equation~\ref{#1}}

\def\1{\bm{1}}

\def\vr{{\bm{r}}}

\def\vx{{\bm{x}}}
\def\vy{{\bm{y}}}

\DeclareMathAlphabet{\mathsfit}{\encodingdefault}{\sfdefault}{m}{sl}
\SetMathAlphabet{\mathsfit}{bold}{\encodingdefault}{\sfdefault}{bx}{n}

\DeclareMathOperator*{\argmax}{arg\,max}
\DeclareMathOperator*{\argmin}{arg\,min}

\usepackage[ruled,boxed,lined]{algorithm2e} 
\let\oldnl\nl
\newcommand{\nonl}{\renewcommand{\nl}{\let\nl\oldnl}}

\title{Amortized Noisy Channel Neural Machine Translation}

\author{Richard Yuanzhe Pang\\
  New York University \\
  \texttt{yzpang@nyu.edu} \\\And
  He He \\
  New York University \\\And
  Kyunghyun Cho \\
  New York University \\
  Genentech\\
  CIFAR Fellow}

\begin{document}
\maketitle
\begin{abstract}
Noisy channel models have been especially effective in neural machine translation (NMT). However, recent approaches like ``beam search and rerank'' (BSR) incur significant computation overhead during inference, making real-world application infeasible. We aim to study if it is possible to build an amortized noisy channel NMT model such that when we do greedy decoding during inference, the translation accuracy matches that of BSR in terms of reward (based on the source-to-target log probability and the target-to-source log probability) and quality (based on BLEU and BLEURT). We attempt three approaches to train the new model: knowledge distillation, 1-step-deviation imitation learning, and Q learning. The first approach obtains the noisy channel signal from a pseudo-corpus, and the latter two approaches aim to optimize toward a noisy-channel MT reward directly. For all three approaches, the generated translations fail to achieve rewards comparable to BSR, but the translation quality approximated by BLEU and BLEURT is similar to the quality of BSR-produced translations. Additionally, all three approaches speed up inference by 1--2 orders of magnitude. 
\end{abstract}

\section{Introduction}
\label{sec:intro}

Noisy channel models have been traditionally used in many tasks, including speech recognition \citep{jelinek1997statistical}, spelling correction \citep{brill-moore-2000-improved}, question answering \citep{echihabi-marcu-2003-noisy}, and statistical machine translation \citep{koehn-etal-2003-statistical}. In machine translation (MT), the probability of the source sentence conditioned on the target-language generation is taken into account when generating a translation. In modern neural machine translation (NMT), the noisy channel approach is successful and often indispensable in many recent top-performing machine translation systems \citep{yee-etal-2019-simple,ng-etal-2019-facebook, chen-etal-2020-facebook,yu-etal-2020-deepmind,tran-etal-2021-facebook}. 

One widely used approach of noisy channel NMT is ``beam search and rerank'' (BSR). Assume a trained forward translator and a trained reverse translator,\footnote{Forward: from the source language to the target language; reverse: from the target language to the source language.} BSR decoding consists of two steps: first, decode using beam search with a large beam size from the forward translation model and store the entire beam; second, rerank the beam using a reward which is the sum of the forward translation log probability and the reverse log probability. 
This approach incurs significant computational overhead, 
given the need to decode a large beam (usually with beam size 50--100) from the forward translator and the need to feed the large beam through the reverse translator. The computational cost is especially problematic if the practitioner has a large volume of translation requests, or if the system is mobile-based and requires offline translation. 

We thus aim to learn a separate neural network with an identical architecture as the forward translator such that at inference time, when we do \textit{greedy decoding} using this new network, we investigate how much translation accuracy would be sacrificed. Specifically, we investigate how forward/reverse rewards of the translations as well as the translation quality (approximated by BLEU and BLEURT) would compare to those of BSR-generated translations.\footnote{Although we need time to train the separate network, at inference time and during the actual large-scale user-facing deployment, we would be able massively cut down computational cost in the long run. In this paper, we aim to investigate the accuracy of such decoded translations.}

The paper explores three approaches, with increasingly more exploration when optimizing the reward. (1) Knowledge distillation (KD) from a pseudo-training-corpus generated by BSR: we can treat the BSR-generated corpus as the oracle, and KD can be interpreted as behavioral cloning. (2) a one-step-deviation imitation learning strategy (IL) where given a fixed sequence of target-language tokens, we adjust each time-step's probability distribution over the vocabulary such that the resulting distribution minimizes an energy function used in BSR reranking, and (3) Q learning which explicitly learns the scoring function used in BSR reranking. 

We experiment on three datasets (IWSLT'14 De-En, WMT'16 Ro-En, and WMT'14 De-En). Experimental results show that all three approaches speed up inference by 50--100 times. The approaches fail to achieve comparable rewards to BSR, but compared to the non-BSR baselines, the approaches achieve much higher reverse rewards (i.e., $\log p_r(\vx \mid \vy)$ where $p_r$ is the reverse translator) at the expense of forward rewards (i.e., $\log p_f(\vy \mid \vx)$ where $p_f$ is the forward translator). Meanwhile, the approaches achieve a translation quality (approximated by BLEU and BLEURT) that is comparable to that of BSR. In particular, IL's BLEURT scores is significantly higher than those of beam search, across all three datasets; IL's BLEURT scores are not significantly different from BSR's scores, across three datasets.

\section{Background}
\label{sec:background}

\subsection{Neural Machine Translation}

NMT systems usually model the distribution $p(\vy \mid \vx)$ where $\vx=(x_1, x_2, \dots, x_{T_s})$ is a source-language sequence and $\vy = (y_1, y_2, \dots, y_T)$ is a target-language sequence. Most NMT systems use an autoregressive factorization: 
$$\log p(\vy \mid \vx) = \sum_{t=1}^T \log p_\theta(y_t \mid \vy_{<t}, \vx),$$   
where $\vy_{<t} = (y_1, y_2, \dots, y_{t-1})$, and $p_\theta$ is parameterized with a neural network. 
At test-time, to decode a translation given a source sentence, greedy decoding and beam search are most commonly used. Both are approximate search methods to find the highest-scoring translations.

\subsection{Beam Search and Rerank (BSR)}

BSR has appeared in a number of top-performing models, including many winning submissions of the WMT competitions 
\citep{ng-etal-2019-facebook, chen-etal-2020-facebook,yu-etal-2020-deepmind,tran-etal-2021-facebook}. The intuition of BSR is to take advantage of the reverse translator during decoding. Specifically, we do beam search with a large beam size $b$ (usually 50--100) to obtain $b$ candidate translations. Then, we rerank the candidates using the scoring function:
\begin{align}
    \log p_f(\vy \mid \vx) + \gamma \log p_r(\vx \mid \vy) + \gamma' \log p_{lm}(\vy), \nonumber
    \label{eq:rerank}
\end{align}
where $\gamma$ and $\gamma'$ are tuned in $[0,2]$. 
Without access to a language model trained on a huge target-language monolingual external corpus, if we use $\log p_f(\vy \mid \vx) + \gamma \log p_r(\vx \mid \vy) $ as the ranking criteria, BSR also provides a significant performance gain. With a large beam size, this approach performs better than the ``two-step beam search'' approach \citep{yu2017neural,yee-etal-2019-simple}.

\section{Amortized Noisy-Channel NMT}
\label{sec:methods}

One common problem with the above approaches is the inference-time computation overhead.  
If a translation system needs to translate a high volume of texts, then the test-time computational efficiency is crucial. 
Thus, our goal is to use a network to approximate such a noisy channel NMT system, while having the same inference-time computational cost as \emph{greedily decoding from $p_f$}. Specifically, we want our translations to maximize the following objective:
\begin{equation}
    R(\vx, \vy) = \log p_f (\vy \mid \vx) + \gamma \log p_r (\vx \mid \vy),  \label{eq:reward}
\end{equation}
where $\gamma > 0$ is some fixed coefficient. {Using the autoregressive factorization, the forward reward $\log p_f (\vy \mid \vx)$ equals $\sum_{t=1}^{|\vy|} \log p_f (\vy_t \mid \vy_{<t}, \vx)$, and the reverse reward $\log p_r (\vx \mid \vy)$ equals $\sum_{t=1}^{|\vx|} \log p_r (\vx_t \mid \vx_{<t}, \vy)$.}

\paragraph{Goal: Investigating if greedily decoding from our new models leads to successful amortization.}
\label{sec:criteria}


Three approaches are shown in this section. We do greedy decoding from the obtained models, and we investigate whether amortization is successful as follows. 
\begin{itemize}
    \item First, we examine if decoding is faster than BSR. This aspect is guaranteed given that we would do greedy decoding from our new network which has the same architecture as $p_f$.
    \item Next, we examine if both the forward and reverse rewards of the translations are close to the forward and reverse rewards of the translations generated by BSR, respectively.
    \item Finally, we examine the translation quality by checking if BLEU and BLEURT scores of our model's translations are close to those of BSR-produced translations. 
\end{itemize}

\subsection{Approach 1: Knowledge Distillation (KD)}

KD has been used to amortize beam search \citep{chen-etal-2018-stable}. It is also effective in NMT in general \citep{kim-rush-2016-sequence,freitag2017ensemble,tan2019multilingual,tu-etal-2020-engine}. Here we adapt a simple version of KD for amortized noisy-channel decoding. 

First, train a forward translator $p_f$ and a reverse translator $p_r$ using maximum likelihood estimation. Then, do BSR on the entire training set to obtain the pseudo-corpus. In particular, we ignore the $p_{lm}$ term in this paper given that it usually requires a big language model, and the inclusion of the term is orthogonal to our goal of reducing inference time.\footnote{Generating the pseudo-corpora can be paralleled. If the system is deployed in the real world, we argue that the amount of computation used to generate the pseudo-corpus is negligible, compared to the aggregate amount of computation for inference.}  

Next, we train a separate ``knowledge distilled'' model $p_{\mathrm{KD}}$ on this new pseudo-corpus (i.e., with the original source-language sentences and the BSR-generated target-language sentences). This objective is equivalent to minimizing the KL-divergence between the distribution induced by the pseudo-corpus obtained through BSR and our model distribution. 

At inference time, we greedily decode from $p_{\mathrm{KD}}$.

\subsection{Approach 2: 1-Step-Deviation Imitation Learning (IL)}
\label{sec:engine}

Define a network $A_\phi$ such that it takes in the source sentence and a target-language prefix, and $A_\phi (\cdot \mid \vx, \vy_{<t})$ outputs a $|\mathcal{V}|$-dimensional probability distribution corresponding to the $t$-th time-step. Moreover, $A_\phi$ and $p_f$ have the same architecture.  
In autoregressive text generation, to learn $A_\phi$ such that it is close to an existing network $p_\theta$, imitation learning seeks to optimize $\phi$ as follows:
\begin{align}
    \argmin_\phi \mathbb{E}_{(\vx, \vy_{<t})} \mathcal{L}(A_{\phi}(\cdot | \vx, \vy_{<t}), p_\theta(\cdot | \vx, \vy_{<t})), \nonumber
\end{align}
where one example of $\mathcal{L}$ is the cross entropy.

\paragraph{Forward energy.}

Inspired by ENGINE \citep{tu-etal-2020-engine}, in the context of noisy channel NMT, define the forward sub-energy $E_t^f$, which is a function of $\phi$, as follows:\footnote{If we compute $p_f(\cdot \mid y_1, y_2, \dots, y_{t-1}, \vx)$ where $y_i$'s correspond to token IDs of a partial translation in the target language, then we would first look up the $y_i$-th row of the embedding matrix $E_{emb}$ and use this row to embed $y_i$; equivalently, we can use the product $\mathrm{onehot}(y_i)^\top E_{emb} $ to embed $y_i$. In the case of $E_t^f$, the prefixes used in $p_f$ are distributions instead of tokens. We can use $A_\phi(\cdot | \vx, \hat{\vy}_{<i})^\top E_{emb}$ to represent the $i$-th token embedding. The embedding strategy is similar for $E_t^r$ below.}
\begin{align}
    & E_t^f (\vx, \hat{\vy}; \phi) = - A_\phi (\cdot \mid \vx, \hat{\vy}_{<t})^\top \nonumber \\ 
    & \quad \log p_f (\cdot \mid A_\phi(\cdot | \vx, \hat{\vy}_{<1}), \dots, A_\phi(\cdot | \vx, \hat{\vy}_{<t}), \vx). \nonumber 
\end{align} 
Suppose we have a source sentence $\vx$ and a sequence of prefix distributions $\hat{\vy}_{<1}, \dots, \hat{\vy}_{<T}$. We call $A_\phi(\cdot \mid \vx, \hat{\vy}_{<t})$ the $t$-th step distribution according to $A_\phi$. Intuitively, given a source and a fixed sequence of prefixes, we learn $A_\phi$ such that the resulting $t$-th-step distribution matches the forward conditional probability (measured by $p_f$) -- the latter depends on the source $\vx$ and the prefix distributions.

\paragraph{Reverse energy.}

Next, we define the reverse sub-energy as follows:
\begin{align}
    & E_t^r (\vx, \hat{\vy}; \phi) = - \mathrm{onehot}(\vx_t)^\top \log p_r (\cdot \mid \vx_{<t}, \nonumber \\ 
    & \qquad A_\phi(\cdot \mid \vx, \hat{\vy}_{<1}), \dots, A_\phi(\cdot \mid \vx, \hat{\vy}_{<T})). \nonumber
\end{align} 
Intuitively, we also learn $A_\phi$ such that the one-hot distributions corresponding to the source words should match the reverse conditional probability (measured by $p_r$). 

\paragraph{Trajectories.}

In the above equations, $\hat{\vy} = (\hat{y}_1, \dots, \hat{y}_T)$. $\hat{y}_t$ comes from two sources, with probability $p$ and $1-p$ for each minibatch during training (Section~\ref{sec:hyperparameters}): 
(i) $\hat{\vy}_t = \argmax_{v \in \mathcal{V}} A_\phi (v \mid \vx, \hat{y}_{<t})$ and $\hat{y}_{<1} = \varnothing$; in other words, given that $A_\phi (\cdot \mid \vx, \hat{y}_{<t})$ is a probability distribution, we use the most likely token as $\hat{y_t}$. 
(ii) For the second source, let $\hat{v}_t$ be the $t$-th token of the BSR-obtained sequence, so that we can expose our model to BSR-prefixes, which are the optimal prefixes.

\paragraph{Final objective.}

We train $A_\phi$ using the following objective:
\begin{align}
    \min_\phi \sum_{\vx} \left[ \sum_{t=1}^{T} E_t^f (\vx, \hat{\vy}; \phi) + \gamma \sum_{t'=1}^{|\vx|}  E_{t'}^r (\vx, \hat{\vy}; \phi) \right].  \nonumber
\end{align} 
During inference, we greedily decode from $A$.

\subsection{Approach 3: Q Learning}
\label{sec:q-learning}

A well-motivated approach is to use Q learning  \citep{watkins1992q,sutton1998reinforcement} to explicitly learn a reward function $Q$, with the goal that when we {greedily} decode from $Q$, the generations maximize the reward shown in Eq.~(\ref{eq:reward}).

Let us view machine translation as a sequential decision-making process. 
At time-step $t$, given a state $s_t = (\vy_{<t}, \vx) $, a policy takes an action $a_t \in \mathcal{V}$,
transits to the next state $s_{t+1} = (\vy_{<(t+1)}, \vx)$,\footnote{$\vy_{<(t+1)}$ equals $\vy_{<t}$ concatenated with the action $a_t$.} and receives a reward $r_t$.

\subsubsection{Background on Q Learning}
\label{app:q-intro}

In Q learning, $Q^\pi: \mathcal{S} \times \mathcal{A} \to \mathbb{R}$ is a function such that $Q^\pi(s_t,a_t)$ produces the expected return after seeing state $s_t$, taking action $a_t$, and following policy $\pi$; i.e., $Q^\pi(s_t,a_t) = \mathbb{E}[ \sum_{t'=t}^\infty r_{t'} | s_t, a_t, \pi ]$ assuming discount factor 1. We further define $Q^*: \mathcal{S} \times \mathcal{A} \to \mathbb{R}$ to be the optimal action-value function: $Q^*(s_t, a_t) = \max_{\pi} \mathbb{E} [ \sum_{t'=t}^\infty r_{t'} | s_t, a_t, \pi]$, which is the maximum return achievable by following any strategy \textit{after} seeing a state $s_t$ and taking an action $a_t$. In particular, $Q^*$ solves the Bellman Equation \citep{sutton1998reinforcement}:
\begin{align}
    Q^*(s_t,a_t) = r_t + \max_{a_{t+1}} Q^*(s_{t+1}, a_{t+1}),
    \nonumber
\end{align}
assuming discount factor 1 and given deterministic transition dynamics (in our machine translation scenario) after taking action $a_t$ given state $s_t$. 

Traditionally, the $Q$ function is implemented as a matrix of size $|\mathcal{S}| \times |\mathcal{A}|$, which is intractable in the case of MT due to the combinatorial nature of the state space. 
We thus use function approximation to tackle this issue of intractability: we follow \citet{mnih2015human} and use a deep neural network trained with experience replay and target networks to approximate the Q learning. 

Deep Q learning draws samples from a set of trajectories $\mathcal{B}$, and the neural network $Q$ aims to predict $Q^*$ by learning based on minimizing the following squared loss. 
\begin{align}
    & L(\phi) = \frac{1}{|\mathcal{B}|} \sum_{(s_t,a_t,s_{t+1},r_t) \sim \text{Uniform}(\mathcal{B})}  \nonumber \\ 
    & \left[( r_t + \max_{a_{t+1}} Q'(s_{t+1},a_{t+1}) - Q(s_t, a_t) )^2 \right], \nonumber
\end{align}
where $\phi$ is the parameter to $Q$, and $Q'$ is a slightly old copy of $Q$.\footnote{In other words, after a fixed number of optimization steps, we update $Q'$ by $Q$.}

\subsubsection{Q Learning for Amortized Noisy Channel MT}

To model the noisy-channel NMT, given a target-language sequence $\vy$ and its length $T$, we have reward $\vr = (r_1, \dots, r_T)$, where 
\begin{align}
r_t = 
    \begin{cases}
    \log p_f(y_t | \vy_{<t}, \vx), & \text{if } t<T, \\
    \log p_f(y_T | \vy_{<T}, \vx) + \\ \qquad \gamma\cdot \log p_r(\vx|\vy), & \text{if } t=T.
    \end{cases}
\label{eq:q-reward}
\end{align}

\begin{algorithm}[t]
{
    \nonl Given $p_f$, $p_r$, and a parallel translation dataset $\mathcal{D}$. \\

     \While{not converged}{
        Collect training trajectories (\S\ref{sec:q-learning}), and sample a mini-batch $\mathcal{B}$. \\
        \medskip
        Compute target $R_t$: 
        \begin{itemize}
            \item if $t<T$, then $R_t = r_t + \max_{a_{t+1}} Q_\phi'(s_{t+1}, a_{t+1})$; 
            \item if $t=T$, then $R_t = r_T$.
        \end{itemize} 

        Update $\phi$ (using gradient descent) by the objective
        $$\argmin_\phi \left[ Q_\phi(s_t, a_t) - R_t \right]^2.$$\\

        Update $Q_\phi'$: $Q_\phi' \leftarrow Q_\phi$ every $K$ steps.\\
    }
}
\caption{Q learning for amortized noisy channel NMT}
\label{algo:q-dqn}
\end{algorithm} 

We construct $Q$ to have the same architecture as $p_f$ without the final softmax layer. $Q$ is trained using Algorithm~\ref{algo:q-dqn} which is adapted from deep Q learning originally applied to Atari games \citep{mnih2015human}, given that we aim to best leverage the existing off-policy trajectories from different sources. The full algorithm is shown in Algorithm~\ref{algo:q-dqn}. 

In short, our algorithm says that given a trajectory $(\vx, \vy, \vr)$, at time-step $t<T$, we want the scalar $Q(s_t, a_t)$ to be close to the sum of the $t$-th step reward 
and the \textit{most optimistic} future return, 
had we taken action $a_t$ at time-step $t$. At time-step $T$, we want $Q(s_T, a_T)=Q(({\vy}_{<T}, \vx), \langle \text{eos} \rangle)$ to be close to $r_T$, as defined in Eq.~(\ref{eq:q-reward}). 

To generate the $t$-th token at {inference-time}, we do greedy decoding using $Q$ as follows: 
$\hat{y}_t = \argmax_{a_t \in \mathcal{V}} Q (s_t, a_t)$.

\paragraph{Trajectories.}

The off-policy algorithm shown in Algorithm~\ref{algo:q-dqn} requires trajectories, i.e., $(\vx, \vy, \vr)$ tuples. The trajectories come from two sources. 

\paragraph{(1) $Q$-based trajectories.} In this category, we have two ways of obtaining $\vy$: (1a) Boltzmann exploration \citep{sutton1990integrated}\footnote{Recall that at time-step $t$, $Q(s_t, a_t) \in \mathbb{R}$ for each $a_t \in V$. Therefore, $Q(s_t, \cdot) \in \mathbb{R^{|\mathcal{V}|}}$. We turn the vector of real numbers to a categorical distribution by softmax with temperature $\gamma_b$. Then, the sequences in the trajectories are obtained by sampling from the aforementioned distribution. In practice, for each sequence, we use a temperature $\gamma_b$ sampled from $\mathrm{Uniform}(0, 1.5)$. One can think of this strategy as a variant of $\epsilon$-greedy which is typically used in Q learning.} and (1b) greedy decoding based on $Q$. 
At the start of the optimization, however, most of the $Q$-generated sequences are very far from target sequences. The lack of high-reward sequences prevents Q learning from efficient optimization. Therefore, we also inject reasonably good trajectories from the beginning of training by utilizing both ground-truth sequences as well as $p_f$-based sequences. 
We thus need the next category of sources. 

\paragraph{(2) $p_f$-based trajectories.} The target-language sequences are obtained by decoding using $p_f$; please find more details in Appendix~\ref{app:trajectories}.\footnote{We have also experimented with gold-standard trajectories from the parallel translation dataset $\mathcal{D}$, but the inclusion of such trajectories do not lead to better rewards of $Q$-generated translations.}

%



\section{Experimental Setup}

\begin{table*}[ht!]
\setlength{\tabcolsep}{2.4pt}
\centering
\small
\begin{tabular}{>{\hangindent=0.9em\hangafter=1}m{3.2cm}ccccccccccc}
    \toprule
    & \multicolumn{3}{c}{IWSLT'14 De-En} & & \multicolumn{3}{c}{WMT'16 Ro-En} & & \multicolumn{3}{c}{WMT'14 De-En} \\
    \cline{2-4} \cline{6-8} \cline{10-12}
    \noalign{\smallskip}
    & \makecell[c]{$b$} & \makecell[c]{fwd reward \\ mean (std)} & \makecell[c]{rvs reward \\ mean (std)} &  & $b$ & \makecell[c]{fwd reward \\ mean (std)} & \makecell[c]{rvs reward \\ mean (std)} & & $b$ & \makecell[c]{fwd reward \\ mean (std)} & \makecell[c]{rvs reward \\ mean (std)}  \\
    \midrule
    $p_f$  & 1 & -9.1 {\scriptsize (7.7)} & -35.4 {\scriptsize (39.9)} & & 1 & -9.5 {\scriptsize (11.5)} & -41.0 {\scriptsize (50.1)} & & 1 & -11.0 {\scriptsize (6.3)} & -31.5 {\scriptsize (24.6)} \\
    $p_f$  & 5  & {-8.6} {\scriptsize (7.0)} & -34.2 {\scriptsize (38.5)} &  & 5 & {-9.0} {\scriptsize (8.5)} & -40.2 {\scriptsize (48.2)} & & 7 & {-10.4} {\scriptsize (5.5)} & -29.9 {\scriptsize (21.5)} \\
    {BSR} & 100 & {-9.4} {\scriptsize (6.8)} & {-25.7} {\scriptsize (32.5)} & & 70 & -10.0 {\scriptsize (6.0)} & {-29.7} {\scriptsize (41.9)} & & 50 & -10.7 {\scriptsize (5.3)} & {-23.6} {\scriptsize (16.3)}\\
    \specialrule{.2pt}{1pt}{1pt}

    KD & {1} & -13.8 {\scriptsize (13.9)} & -28.0 {\scriptsize (32.7)} & & 1 & -17.2 {\scriptsize (26.3)} & -35.4 {\scriptsize (44.6)} & & 1  & -14.8 {\scriptsize (9.1)} & -24.0 {\scriptsize (16.7)} \\ 
    IL & {1} &  -13.3 {\scriptsize (13.2)} & -27.9 {\scriptsize (32.3)} &  & {1} &  -17.2 {\scriptsize (30.9)} & -34.3 {\scriptsize (45.3)} & & {1} & -14.6 {\scriptsize (8.9)} & {-23.6} {\scriptsize (15.9)}  \\ 
    Q learning & {1} & -13.7 {\scriptsize (21.4)} & -29.9 {\scriptsize (35.1)} & & 1 & -11.6 {\scriptsize (19.7)} & -39.1 {\scriptsize (52.9)} & & {1} & -14.4 {\scriptsize (9.9)} & -24.9 {\scriptsize (17.5)} \\  
    \specialrule{.2pt}{1pt}{1pt}

    reference data & -- & -38.8 {\scriptsize (39.7)} & -45.2 {\scriptsize (46.6)} & & -- &  -55.3 {\scriptsize (51.1)} & -59.0 {\scriptsize (54.2)} & & -- & -36.8 {\scriptsize (24.4)} & -36.8 {\scriptsize (23.0)}  \\
\bottomrule
\end{tabular}
\caption{Mean and standard deviation (across sequences) of test set forward and reverse rewards for translations. $b$ refers to beam size during inference. 
}
\label{tab:main}
\end{table*}

\begin{table}[ht!]
\setlength{\tabcolsep}{2.1pt}
\centering
\small
\begin{tabular}{>{\hangindent=0.9em\hangafter=1}m{2.8cm}cccccccccc}
    \toprule
    & \makecell[c]{IWSLT'14 \\ De-En} & \makecell[c]{WMT'16 \\ Ro-En} & \makecell[c]{WMT'14 \\ De-En}   \\
    \midrule
    $p_f$ (greedy decoding)  & 33.65 {\scriptsize (0.06)} & 33.23 {\scriptsize (0.14)} & 30.39 {\scriptsize (0.13)}  \\
    $p_f$ (beam search) & 34.54 {\scriptsize (0.08)} & 33.98 {\scriptsize (0.15)} & 31.78 {\scriptsize (0.08)} \\
    {BSR} & {35.43} {\scriptsize (0.06)} & {34.81} {\scriptsize (0.09)} & {32.15} {\scriptsize (0.14)}   \\
    \specialrule{.2pt}{1pt}{1pt}

    KD & 35.39 {\scriptsize (0.04)} & 33.95 {\scriptsize (0.10)} & 31.71 {\scriptsize (0.05)}\\ 
    IL & {35.61} {\scriptsize (0.09)}  & {34.65} {\scriptsize (0.07)} & 31.90 {\scriptsize (0.07)}  \\ 
    Q learning & 34.60 {\scriptsize (0.08)} & 34.31 {\scriptsize (0.15)} & 31.60 {\scriptsize (0.19)}  \\  
\bottomrule
\end{tabular}
\caption{Test set sacreBLEU (mean \& standard deviation of three runs using different random seeds). IL performs the best among the three proposed methods. 
}
\label{tab:main-bleu}
\end{table}

\begin{table}[ht!]
\setlength{\tabcolsep}{2.1pt}
\centering
\small
\begin{tabular}{>{\hangindent=0.9em\hangafter=1}m{2.8cm}cccccccccc}
    \toprule
    & \makecell[c]{IWSLT'14 \\ De-En} & \makecell[c]{WMT'16 \\ Ro-En} & \makecell[c]{WMT'14 \\ De-En}   \\
    \midrule
    $p_f$ (greedy decoding)  & 62.40 {\scriptsize (0.04)} & 61.14 {\scriptsize (0.10)} & 64.83 {\scriptsize (0.10)}  \\
    $p_f$ (beam search) & 63.21 {\scriptsize (0.07)} & 61.42 {\scriptsize (0.15)} & 65.79 {\scriptsize (0.08)} \\
    {BSR} & {64.15} {\scriptsize (0.05)} & {62.67} {\scriptsize (0.13)} & {66.32} {\scriptsize (0.12)}   \\
    \specialrule{.2pt}{1pt}{1pt}

    KD & 63.88 {\scriptsize (0.04)} & 61.78 {\scriptsize (0.10)} & 66.00 {\scriptsize (0.07)}\\ 
    IL & {63.94} {\scriptsize (0.13)}  & {62.35} {\scriptsize (0.16)} & 66.14 {\scriptsize (0.08)}  \\ 
    Q learning & 63.25 {\scriptsize (0.07)} & 61.70 {\scriptsize (0.18)} & 65.92 {\scriptsize (0.14)}  \\  
\bottomrule
\end{tabular}
\caption{Test set BLEURT-20-D12 (mean \& standard deviation of three runs). IL performs the best among the three proposed methods. Significance test is conducted in Table~\ref{tab:main-bleurt-sigtest}, which shows that IL's scores are significantly better than the scores by beam search; in addition, IL's scores are not significantly different from BSR's scores. 
}
\label{tab:main-bleurt}
\end{table}

\subsection{Tasks and Models}

We experiment on three translation tasks: IWSLT 2014 German to English \citep[IWSLT'14 De-En;][]{cettolo2014report} which has a small training set (train/dev/test size: 160,239/7,283/6,750), WMT 2016 Romanian to English \citep[WMT'16 Ro-En;][]{bojar-etal-2016-findings} which has a medium-sized training set (train/dev/test size: 608,319/1,999/1,999), and WMT 2014 German to English \citep[WMT'14 De-En;][]{bojar-etal-2014-findings} which has a moderately large training set (train/dev/test size: 4,500,966/3,000/3,003). 
Each of the transformer models (the $p_{\mathrm{KD}}$ in KD, the $A$ in IL, the $Q$ function in Q learning) has the same number of parameters as the original MLE-trained forward translator $p_f$. The model for IWSLT'14 De-En is the smallest, and the model for WMT'14 De-En is the largest. The detailed settings can be found in Appendix~\ref{app:experiment}. BLEU scores in this paper are computed with sacreBLEU \citep{post-2018-call}. BLEURT scores are computed using BLEURT-20-D12 \citep{sellam-etal-2020-bleurt}, a recent RemBERT-based checkpoint that achieves high human agreement.
The models we experiment on are shown in Table~\ref{tab:main}. 

\subsection{Hyperparameters}
\label{sec:hyperparameters}

The architecture and optimization details of $p_f$ and $p_r$ are shown in Appendix~\ref{app:experiment}. When training $p_f$ and $p_r$, we validate the model performance after each epoch, and select the model that corresponds to the best dev set BLEU. 

$\gamma$ is the coefficient multiplied to the reverse reward, when computing the total reward in Eq.~(\ref{eq:reward}); $\gamma$ and BSR beam size $b$ are tuned on dev set BLEU using BSR. We choose $\gamma=0.9$ and $b=100$ for IWSLT'14 De-En; $\gamma=0.5$ and $b=70$ for WMT'16 Ro-En; $\gamma=0.5$ and $b=50$ WMT'14 De-En. See Appendix~\ref{app:experiment} for details.

For training the IL-based network, the learning rate is selected from $\{10^{-6}, 5\times 10^{-6}, 10^{-5}, 3\times 10^{-5}, 5\times 10^{-5}\}$. We use weight decay of $10^{-4}$. Dropout rate is selected from $\{0, 0.05, 0.1, 0.3\}$; we find that a dropout rate of 0 or 0.05 always works the best. We use a fixed max batch length (i.e., the max number of input tokens in a batch) of 4,096 tokens. The probability $p$, described in Section~\ref{sec:methods}, is selected from $\{0,0.1,0.5,0.9,1\}$; we find that $p=0.1$ or $p=0.5$ usually works the best. We accumulate gradients and do gradient descent once every $k$ steps for computational reasons. 
$k$ is selected from $\{4, 8, 16\}$. We find that the IL approach relies on a good initialization, so we use $p_{\mathrm{KD/nc}}$ to initialize the new network.

For Q learning, the synchronization frequency $K$ in Algorithm~\ref{algo:q-dqn} is selected from $\{10, 20, 30, 50, 150\}$. The learning rate is tuned in $\{10^{-5}, 3\times 10^{-5}, 5\times 10^{-5}, 10^{-4}\}$. We use weight decay of $10^{-4}$. Dropout rate is tuned in $\{0, 0.01, 0.05, 0.1\}$; we find that a dropout rate of 0 always works the best. We use a fixed max batch length 4096. We tune the number of steps per gradient update in $\{4, 8, 16\}$; a large number effectively increases the batch size. The ratio for different trajectories is described in Appendix~\ref{app:trajectories}. Furthermore, we find that training $Q$ with a small $\gamma$ at the beginning stabilizes the training, so we first use $\gamma=0.1$ and train till convergence, and then increase $\gamma$ by 0.2 increment, and we reiterate the process until reaching the desired $\gamma$. 

We use the Adam optimizer \citep{kingma2014adam} for all experiments. We cap the maximum length of the translation at $1.2 T_s + 20$ during decoding, where $T_s$ is the length of a source sentence. 
All implementation is based on fairseq \citep{ott-etal-2019-fairseq}. Each experiment is run on one NVIDIA RTX 8000 GPU. 

\section{Results}

\subsection{Preliminary Analysis}


\paragraph{Inference speed.}

Using any of the three proposed approaches achieves a significant speedup, given that the three approaches all use greedy decoding. We quantify this speedup experimentally. During inference, we maximize the memory usage of a single NVIDIA RTX 8000 GPU by finding the largest batch length in the form of $2^k$ where $k$ is a positive integer.\footnote{Batch length means the number of tokens in a batch (instead of the number of sequences).} 
In the IWSLT'14 De-En task, the inference speed (sequences per second) for BSR is 11. The speed for ``greedy by $p_f$'' is around 1050, and the decoding speed for any of three proposed approaches is also similar.

\paragraph{Rewards.} 

First, comparing the three approaches to greedy decoding or beam search from $p_f$, we see that the three approaches achieve smaller forward rewards, but much larger reverse rewards. This observation is expected given that the three approaches consider both the forward and reverse rewards, while greedy decoding or beam search from $p_f$ only consider forward rewards.   
Second, comparing the three approaches against BSR, the three approaches achieve both smaller forward rewards and smaller reverse rewards. 
However, we find this a reasonable trade-off to be made between decoding latency and rewards, as all these approaches are 1--2 orders of magnitude faster in decoding. 

Among the three approaches, KD and IL achieve a better balance between forward and reverse rewards.  
This observation can be explained by the difference in how the reverse reward is presented among the three approaches. 
In KD and IL, the learning signal by reverse rewards is implicitly spread throughout all the steps in a sequence. In other words, changing the conditional distribution in each time-step would adjust the loss in KD and the reverse energies in IL. In Q learning, the reverse reward is sparse: it only appears at the end of the sequence, unlike the forward reward which is spread throughout all the steps. This makes it easier for Q learning to maximize the forward reward compared to the reverse reward which requires many more updates to be propagated toward the earlier time steps.

\paragraph{Translation quality.} 

The three approaches achieve BLEU and BLEURT scores that are comparable to those by BSR. Moreover, the three approaches achieve BLEU scores that are much better than ``greedy decoding from $p_f$'' which has the same computational budget; they are often better than ``beam search from $p_f$'' as well. In particular, Table~\ref{tab:main-bleurt} shows that IL's BLERUT scores are significantly higher than the scores of beam search across all three datasets. In addition, IL's BLEURT scores are not significantly different from BSR across all three datasets. Therefore, our approaches are able to generate translations with similar quality as those by BSR, while being 5--7 times as fast as beam search and 50--100 times as fast as BSR.

\subsection{Analysis of Translations}
\label{sec:analysis-translations}

\begin{table*}[ht!]
\setlength{\tabcolsep}{1.7pt}
\centering
\small
\begin{tabular}{>{\hangindent=0.9em\hangafter=1}m{2.3cm}cccccccccccccc}
    \toprule
    & \multicolumn{4}{c}{IWSLT'14 De-En} & & \multicolumn{4}{c}{WMT'16 Ro-En} & & \multicolumn{4}{c}{WMT'14 De-En} \\
    \cline{2-5} \cline{7-10} \cline{12-15}
    \noalign{\smallskip}
    & \makecell[c]{$b$} & \makecell[c]{fwd reward \\ mean (std)} & \makecell[c]{rvs reward \\ mean (std)} & BLEU &  & $b$ & \makecell[c]{fwd reward \\ mean (std)} & \makecell[c]{rvs reward \\ mean (std)} & BLEU & & $b$ & \makecell[c]{fwd reward \\ mean (std)} & \makecell[c]{rvs reward \\ mean (std)} & BLEU \\
    \midrule

    $p_{\mathrm{KD}/\mathrm{beam}}$ trained by $(X, \widetilde{Y}_{\mathrm{beam})}$ & {1}  & -13.3 {\scriptsize (13.4)} & -31.6 {\scriptsize (35.2)} & 34.80 & & 1  & -17.0 {\scriptsize (17.4)} & -38.9 {\scriptsize (49.0)}  & 33.22 & & 1 & -14.7 {\scriptsize (9.2)} & -28.0 {\scriptsize (19.3)} & 31.38 \\ 

    $p_{\mathrm{KD}/\mathrm{nc}}$ trained by $(X, \widetilde{Y}_{\mathrm{NC}})$  & {1} & -13.8 {\scriptsize (13.9)} & -28.0 {\scriptsize (32.7)} & 35.39 & & 1 & -17.2 {\scriptsize (26.3)} & -35.4 {\scriptsize (44.6)} & 33.95 & & 1  & -14.8 {\scriptsize (9.1)} & -24.0 {\scriptsize (16.7)} & 31.71 \\ 

\bottomrule
\end{tabular}
\caption{The rewards and BLEU scores using two KD approaches: $p_{\mathrm{KD}/\mathrm{beam}}$ uses the pseudo-corpus generated by doing beam search from $p_f$. $p_{\mathrm{KD}/\mathrm{nc}}$ uses the pseudo-corpus generated by BSR. 
}
\label{tab:kd}
\end{table*}

In Q learning, the reverse reward is only presented as a learning signal at the end of each sequence. As observed earlier by \citet{welleck-etal-2020-consistency}, the length of the generations may inform us of the possible degeneracies, such as excessive repetitions. 

\begin{figure}[t!]
     \centering
         \includegraphics[width=1.01\columnwidth]{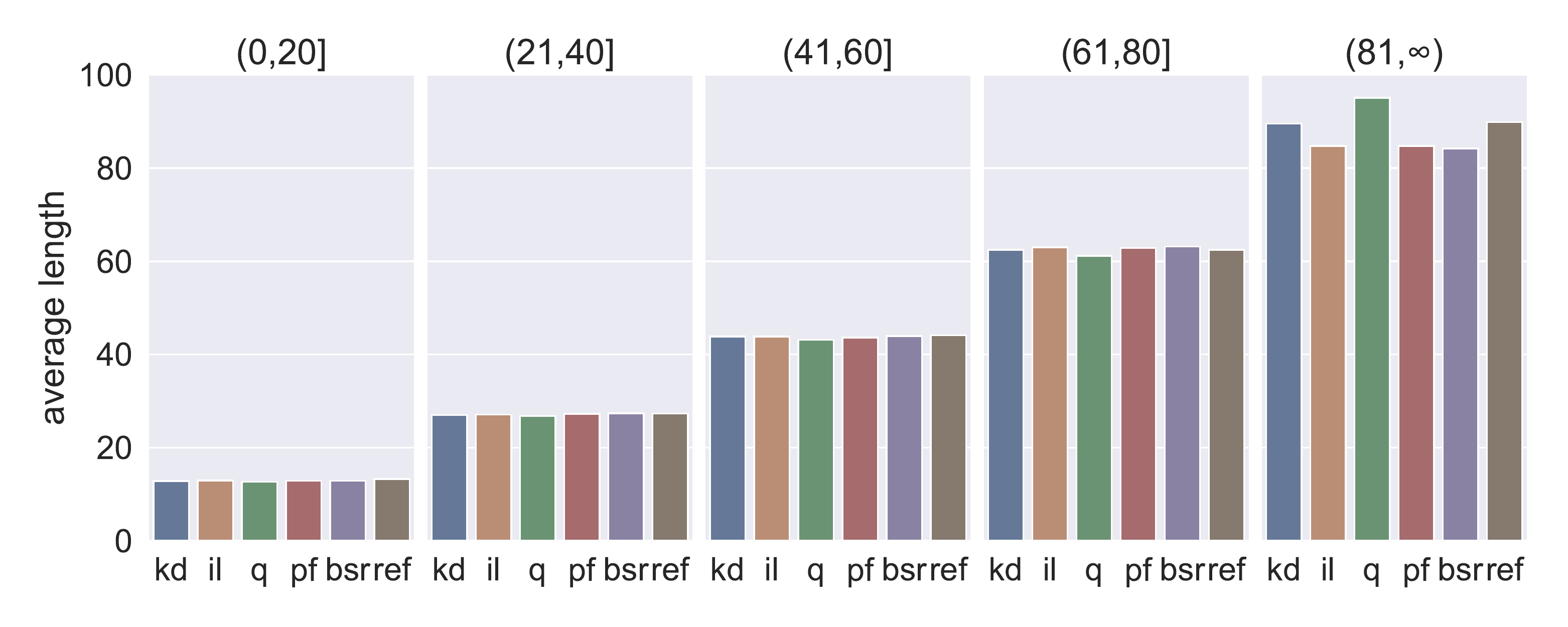}
    \caption{Average length bucketed by length of the source sentence. The five buckets contain 453, 877, 376, 92, 26 sentences, respectively. The six systems are KD, IL, Q learning, beam search by $p_f$, BSR, and reference translations, respectively. In the longest length bucket, Q learning produces translations that are longer than translations by other systems.}
    \label{fig:bucket_len}
\end{figure}

Therefore, we analyze WMT'16 Ro-En translations generated by different systems, and we first examine the lengths of translations in different source length buckets. Figure~\ref{fig:bucket_len} shows that the lengths by different systems are similar in the first four buckets, but in the longest source length bucket $(81,\infty)$, Q learning produces longer translations. 

Closer examination of the translations reveal that Q learning produces degenerate translations with extensive repetitions when the source sentences are among the longest in the entire dev set; other models do not have this issue. Some randomly selected examples are shown in Table~\ref{tab:examples}.

To confirm this finding, we analyze repetitions by source-length buckets. 
We define ``token rep'' to be the percentage of tokens that have appeared in the immediately preceding $5$-grams:
\begin{align}
    \frac{\sum_{i=1}^N \sum_{t=6}^{T^{(i)}} \mathbbm{1}\left[y_{t}^{(i)} \in \{y_{t-5}^{(i)},\dots,y_{t-1}^{(i)}\} \right] }{\sum_{i=1}^N \sum_{t=6}^{T^{(i)}} 1}, \nonumber
\end{align}
where the superscript indicates the $i$-th example, and $N$ indicates the number of translations. 

\begin{figure}[t!]
     \centering
         \includegraphics[width=1.02\columnwidth]{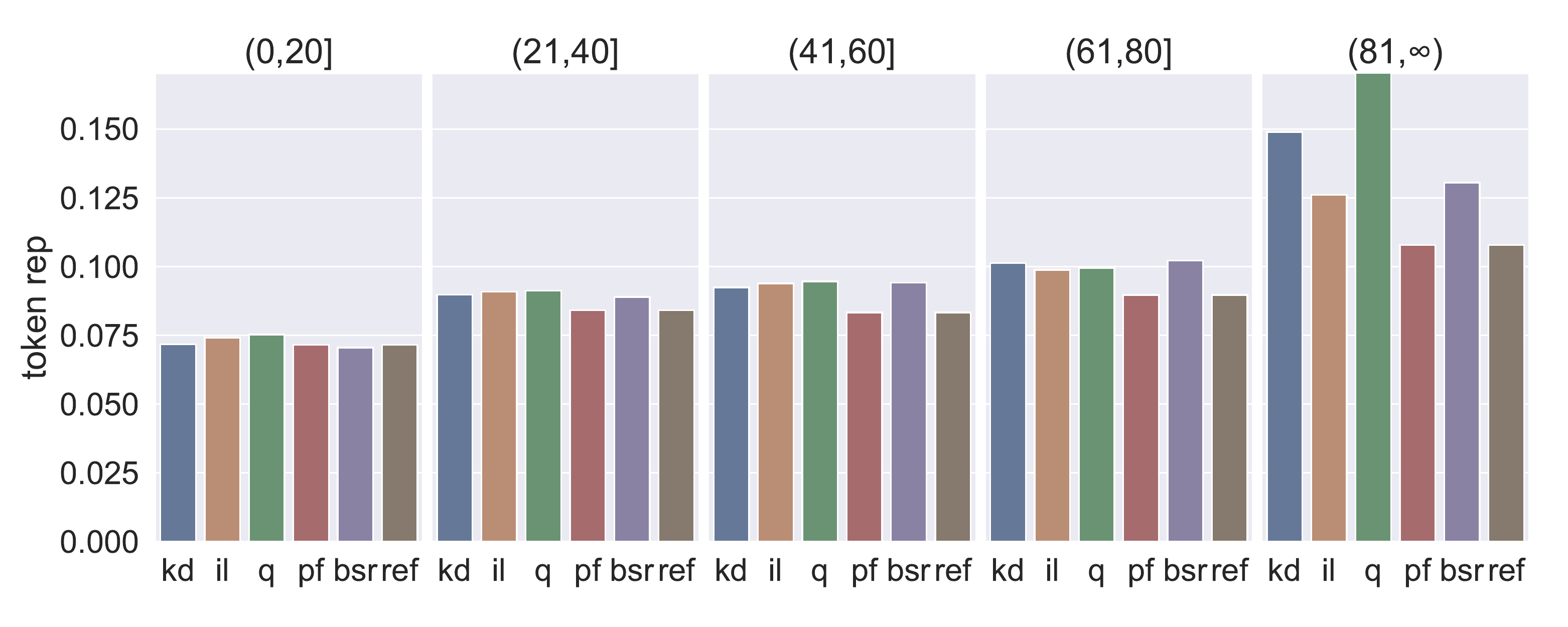}
    \caption{Repetition rate (``token rep'') bucketed by length of the source sentence. The five buckets contain 453, 877, 376, 92, 26 sentences, respectively. N.B.: In the last bucket, ``token rep'' for $Q$-generated translations is around 0.31, and the bar is truncated.}
    \label{fig:bucket_rep}
\end{figure}

We see from Figure~\ref{fig:bucket_rep} that for the longest source-sentence length bucket $(81, \infty)$, $Q$ produces translations with a significantly larger $5$-gram repetition rate. 
Moreover, beam search from the forward only model $p_f$ exhibits a behavior most similar to reference translations. We leave it for the future to study the cause behind an elevated level of repetition in noisy-channel decoding.

\begin{table}[ht!]
\setlength{\tabcolsep}{1.5pt}
\centering
\small
\begin{tabular}{>{\hangindent=1.0em\hangafter=1}m{2.8cm}ccccc}
    \toprule
    & \multicolumn{4}{c}{system 2} \\
    \cline{2-6} 
    \noalign{\smallskip}
    system 1 & \makecell[c]{$p_f$ (beam \\ search)} & BSR & KD & IL & \makecell[c]{Q \\ learning} \\
    \midrule
    $p_f$ (beam search) & 100 & -- & -- & -- & -- \\
    BSR & 81.2 & 100 & -- & -- & -- \\
    KD & 64.5 & 66.0 & 100 & -- & -- \\
    IL & 64.4 & 66.2 & 70.8 & 100 & -- \\
    Q learning & 74.0 & 72.0 & 64.3 & 64.1 & 100  \\  
\bottomrule
\end{tabular}
\caption{Corpus-level BLEU between translations by pairs of systems. Each reported BLEU is averaged between two directions.  
}
\label{tab:pairwise-bleu}
\end{table}

Next, to compare translation similarity among different approaches, we examine the corpus-level BLEU score between each pair of approaches, averaged between two directions. By Table~\ref{tab:pairwise-bleu}, translations by BSR is similar to those produced by $p_f$ and Q learning, compared to KD and IL. 
Now we compare the translations produced by the three approaches. Translations by KD are more similar to IL, compared to BSR and Q learning. This is in line with our intuition that KD and IL differ from Q learning, given that how the reverse reward is presented is different between KD/IL and Q learning.

\begin{table*}[h!]
\scriptsize
\begin{tabular}{p{15.5cm}}
\toprule
\textbf{source:} acum , insa , tsipras cere grecilor sa ii incredinteze din nou mandatul de premier , in cadrul unor alegeri despre care sustine ca ii vor intari pozitia politica . \\  
\textbf{KD:} now , however , tsipras is urging greeks to entrust the prime minister \&apos;s mandate again , in an election he claims will strengthen his political position . \\ 
\textbf{IL:} now , however , tsipras is asking greeks to reentrust them with the prime minister \&apos;s term , in an election that they claim will strengthen his political position . \\ 
\textbf{Q learning:} now , however , tsipras is urging greeks to reentrust his term as prime minister in an election that he claims will strengthen his political position . \\
\textbf{beam search by $p_f$}: now , however , tsipras is urging greeks to re-entrust the prime minister \&apos;s term in an election that he claims will strengthen his political position . \\
\textbf{BSR:} now , however , tsipras is urging greeks to reentrust the prime minister \&apos;s term , in an election that he claims will strengthen his political stance . \\ 
\textbf{reference:} now , however , tsipras asks the greeks again to entrust him with the prime minister position , during an election which he says will strengthen his political position . \\

\midrule 
\textbf{source:} adomnitei a fost trimis in judecata de directia nationala antico <unk> ruptie ( dna ) , fiind acuzat de favorizarea faptuitorului si fals intelectual dupa ce , spun pro <unk> curorii , ar fi incercat sa mascheze un control de audit in urma caruia se descoperise o serie de nereguli cu privire la receptia dintr-un contranct public semnat intre cj si firma laser co . \\  
\textbf{KD:} adomnitei was sued by the national anti-co nistelrooij ruptie ( dna ) as accused of favouring the perpetrator and false intellectual after , pro nistelrooij curorii says , he would have tried to disguise an audit control as a result of which a number of irregularities concerning reception in a public contranct signed between the cj and laser co were discovered . \\ 
\textbf{IL:} adomnitei was sued by the national directorate antico iel ruptie ( dna ) and accused of favouring the perpetrator and forgery an intellectual after , pro iel curorii says , he had tried to disguise an audit control that found a number of irregularities regarding the reception in a public conctrant signed between cj and laser . \\ 
\textbf{Q learning:} the runner the runner , the runner-the runner-in-ranging runner-up is given to the latter , as he is accused of promoting the perpetrator and faltering intellectual after , says pro or: curors , tried to disguise an audit control , as a result of which a number of irregularities concerning a reception signed between cj and laser had been discovered in a public cross-border convoy . \\
\textbf{beam search by $p_f$:} adomnitei was sued by the national anti-co nistelrooij rupture ( dna , accused of favouring the perpetrator and forgery intellectual after allegedly attempting to disguise an audit control line between cj and lasco . \\
\textbf{BSR:} adomnitei was sued by the national anti-co xiated department ( dna , accused of favouring the perpetrator and forgery intellectual after allegedly attempting to disguise an audit control line signed between cj and the lasco firm . \\ 
\textbf{reference:} adomni<<unk>> ei was indicted by the national anticorruption directorate ( dna ) , being accused of favouring the offender and forgery after , according to the prosecutors , he tried to mask an audit which discovered a number of irregularities regarding the acceptance of a public contract entered into by the county council and the company laser co . \\
\bottomrule
\end{tabular}
\caption{WMT'16 Ro-En examples produced by different systems. The top example is randomly selected. The bottom example is an example with a long source, and Q learning produces repetitions. 
\label{tab:examples}
}
\end{table*}

\subsection{Further Analysis}
\label{sec:discussion}

\paragraph{KD.} 

One may wonder whether the improvements in KD arise from the KD procedure or because we use BSR when constructing the pseudo-corpus. We therefore experiment with another model $p_{\mathrm{KD}/\mathrm{beam}}$: we generate the pseudo-corpus $\widetilde{Y}_{\mathrm{beam}}$ from the training set, by beam search from $p_f$, and then use MLE to train $p_{\mathrm{KD}/\mathrm{beam}}$ using the parallel corpora $(X, \widetilde{Y}_{\mathrm{beam}})$. Table~\ref{tab:kd} suggests that the forward rewards of the two approaches are similar, but the reverse rewards for $p_{\mathrm{KD}/\mathrm{nc}}$ is much larger. Meanwhile, $p_{\mathrm{KD}/\mathrm{nc}}$ produces translations with higher BLEU. It is therefore necessary to use BSR  to generate the pseudo-corpus, in order to amortize noisy-channel NMT using KD.

\paragraph{Q learning.}

Why does Q learning, the best understood approach among the three, fail to achieve rewards that are comparable to BSR? The two challenges of a general deep Q learning algorithm are {exploration} and {optimization}. 

Exploration refers to whether we can find high-quality trajectories. We hypothesize that it is not an issue given the diversity of trajectories we use, as shown in  Appendix~\ref{app:trajectories}. We even attempt adding high-reward trajectories from BSR as well as trajectories from a deep ensemble of multiple $p_f$'s but neither BLEU nor reward improves.  

We thus suspect optimization as a challenge. The reverse reward $\log p_r(\vx|\vy)$ is  \textit{sparse} in that it is non-zero only at the terminal state $(\vy_{1:T}, \vx)$ where $y_T = \langle \text{eos} \rangle$.  
The difficulty in maximizing the sparse reverse reward comes from using one-step bootstrapping in Q learning. Such bootstrapping allows Q learning to cope with very long episodes or even an infinite horizon, but this slows down the propagation of future reward to the past. Because we always work with relatively short episodes only in machine translation, we should investigate other learning paradigms from reinforcement learning, such as R learning \citep{mahadevan1996average}. We leave this further investigation to the future.

\section{Related Work}
\label{sec:related}

One of our approaches adapts knowledge distillation (KD) for the noisy channel NMT setting. KD \citep{hinton2015distilling, kim-rush-2016-sequence} has been shown to work well for sequence generation.  
\citet{chen-etal-2018-stable} propose trainable greedy decoding, in which they use knowledge distillation to train a greedy decoder so as to amortize the cost of beam search. More subsequent studies have demonstrated the effectiveness of KD in neural machine translation \citep{freitag2017ensemble,tan2019multilingual}; \citet{gu-etal-2017-trainable} show that it is difficult for on-policy reinforcement learning (RL) to work better than KD. Recently, KD has greatly boosted performance of non-autoregressive MT models \citep{gu2018nonautoregressive,lee-etal-2018-deterministic,tu-etal-2020-engine}. KD is also used to speed up speech synthesis and the approach has been widely deployed in real products \citep{oord2018wavenet}. 

RL for sequence generation has been greatly inspired by \citet{sutton1998reinforcement}. \citet{ranzato2016sequence} and \citet{bahdanau2016actor} apply on-policy RL (REINFORCE and actor-critic algorithms) to MT, but the major optimization challenge lingers given that the reward is usually sparse. \citet{Choshen2020On} recently find that the improvements in MT performance may rely on a good initialization. To address the sparsity issue, \citet{norouzi2016raml} attempt a hybrid maximum likehood (ML) and RL approach. More recently, \citet{pang2021text} attempt to use an offline RL setting with per-token reward based on the a translator trained using standard MLE. 

In recent years, off-policy RL methods have been used to better leverage existing trajectories in text generation. For instance, in the chatbot setting \cite{serban2017deep,zhou2017end}, the periodically-collected human feedback is treated as the trajectory. In our case, we 
leverage the expensive BSR-obtained trajectories as well as trajectories from many different models and sources, although the sparse reward issue still lingers.  

Finally, we point out a recent endeavor to speed up noisy channel NMT inference \citep{bhosale-etal-2020-language}. They reduce the size of the channel model, the size of the output vocabulary, and the number of candidates during beam search. Our solution is orthogonal: we aim to use a separate network to amortize decoding cost, while not changing the network's architecture.

\section{Conclusion}

We describe three approaches (KD, IL, Q learning) to train an amortized noisy-channel NMT model. We investigate whether greedily decoding from these models will lead to accurate translations in terms of reward and quality. Although all three approaches fail to achieve comparable rewards to BSR, the reverse rewards are much higher than those from non-BSR baselines, often at the expense of forward rewards. 
However, we found the translation quality (measured by BLEU and BLEURT) to be comparable to that of BSR, while inference is much faster. 
For future work, the research community could further investigate better ways to optimize toward a sparse reward in the language generation context. Another way to approach the Q learning optimization challenge is to find better reward functions including denser rewards.

\section*{Acknowledgement}

We thank Eneko Agirre, Jon Ander Campos, Kevin Gimpel, Nitish Joshi, Elman Mansimov, and Ethan Perez (alphabetical order) for valuable discussion. This work is supported by Samsung Advanced Institute of Technology (under the project \textit{Next Generation Deep Learning: From Pattern Recognition to AI}) and NSF Award 1922658 NRT-HDR: FUTURE Foundations, Translation, and Responsibility for Data Science.

\bibliography{anthology,custom}
\bibliographystyle{acl_natbib}

\clearpage

\appendix

\section{More Information on Q learning for Amortized Noisy Channel NMT}

\subsection{Details on trajectories}
\label{app:trajectories}

We have obtained trajectories from different sources in the off-policy algorithm (Algorithm~\ref{algo:q-dqn}). Each trajectory contains a source-language sequence $\vx$, a target-language sequence $\vy$, and the corresponding sequence of rewards $\vr = (r_1, \dots, r_T)$. 

One natural category of trajectories to consider is the ones obtained by $Q$ during training. Source (1a) and source (1b) correspond to $Q$-based trajectories. 

Source (2) corresponds to $p_f$-obtained trajectories. Specifically, we split this category into a few sub-sources. (2a) The $\vy$ is obtained through sampling from $p_f$ with temperature sampled from $\mathrm{Uniform}([0,1])$. (2b) The $\vy$ is obtained through greedily decoding from $p_f$. (2c) The $\vy$ is obtained through beam search from $p_f$ with a beam size randomly chosen from 2 to 10. (2d) The $\vy$ is obtained through beam search from $p_f$: we first obtain 50 candidate sequences corresponding to largest $p_f$ probabilities using beam search with beam size 50; next, we pick a random sequence out of these 50 sentences.  

We have also experimented with gold-standard trajectories from the parallel translation dataset $\mathcal{D}$, but the inclusion of such trajectories do not lead to better rewards (of translations generated from $Q$).

The probability for using (1a), (1b), (2a), (2b), (2c), (2d) sequences are 0.3, 0.2, 0.2, 0.1, 0.1, 0.1, respectively.

\begin{table*}[ht!]
\centering
\small
\begin{tabularx}{\linewidth}{Xccccc}
    \toprule
    & \multicolumn{1}{c}{IWSLT'14 De-En} & & \multicolumn{1}{c}{WMT'16 Ro-En} & & \multicolumn{1}{c}{WMT'14 De-En} \\
    \midrule
    \makecell[X]{encoder embedding dimension} & 512 & & 512 & & 1,024 \\
    \makecell[X]{number of encoder attention heads} & 4 & & 8 & & 16 \\
    \makecell[X]{encoder ffn embedding dimension} & 1,024 & & 2,048 & & 4,096 \\
    \makecell[X]{encoder layers} & 6 & & 6 & & 8 \\
    
    \specialrule{.2pt}{1pt}{1pt}
    
    \makecell[X]{decoder embedding dimension} & 512 & & 512 & & 1,024 \\
    \makecell[X]{number of decoder attention heads} & 4 & & 8 & & 16 \\
    \makecell[X]{decoder ffn embedding dimension} & 1,024 & & 2,048 & & 4,096 \\
    \makecell[X]{decoder layers} & 6 & & 6 & & 8 \\
    
    \specialrule{.2pt}{1pt}{1pt}

    \makecell[X]{learning rate} & 0.0005 & & 0.0005 & & 0.0005 \\
    \makecell[X]{dropout rate} & 0.3 & & 0.1 & & 0.1 \\
    \makecell[X]{\# tokens in a batch} & 4,096 ($2^{12}$) & & 65,536 ($2^{16}$) & & 65,536 ($2^{16}$) \\

\bottomrule
\end{tabularx}
\caption{Settings for the forward model $p_f$ and the reverse (channel) model $p_r$.
}
\label{tab:pf-details}
\end{table*}

\begin{table}[ht!]
\setlength{\tabcolsep}{2.1pt}
\centering
\small
\begin{tabular}{>{\hangindent=0.4em\hangafter=1}m{2cm}cccccccccc}
    \toprule
    & \makecell[c]{IWSLT'14 \\ De-En} & \makecell[c]{WMT'16 \\ Ro-En} & \makecell[c]{WMT'14 \\ De-En}   \\
    \midrule
    $p_f$ (greedy)  & 62.40 {\scriptsize (0.04)} & 61.14 {\scriptsize (0.10)} & 64.83 {\scriptsize (0.10)}  \\
    $p_f$ (beam) & 63.21 {\scriptsize (0.07)} & 61.42 {\scriptsize (0.15)} & 65.79 {\scriptsize (0.08)} \\
    {BSR} & {64.15} {\scriptsize (0.05)} & {62.67} {\scriptsize (0.13)} & {66.32} {\scriptsize (0.12)}   \\
    \specialrule{.2pt}{1pt}{1pt}

    KD & 63.88 {\scriptsize (0.04) $*$} & 61.78 {\scriptsize (0.10) $*$} & 66.00 {\scriptsize (0.07) $*$}\\ 
    IL & {63.94} {\scriptsize (0.13) $*\dagger$}  & {62.35} {\scriptsize (0.16) $*\dagger$} & 66.14 {\scriptsize (0.08) $*\dagger$}  \\ 
    Q learning & 63.25 {\scriptsize (0.07)} & 61.70 {\scriptsize (0.18)} & 65.92 {\scriptsize (0.14)}  \\  
\bottomrule
\end{tabular}
\caption{Test set BLEURT-20-D12 (mean \& standard deviation of three runs). IL performs the best among the three proposed methods.  $*$: The score is significant (p-value smaller than 0.05) compared to the beam search results. $\dagger$: The score is significantly higher (p-value smaller than 0.05) than BSR results, or the score is not significantly different (p-value larger than 0.05) from the BSR results.
}
\label{tab:main-bleurt-sigtest}
\end{table}

\section{More Discussion on Experiments}
\label{app:experiment}

\paragraph{BSR hyperparameters.} 
$\gamma$ is tuned in $\{0.1,0.3,0.5,0.7,0.9,1.1,1.3,1.5\}$, and $b$ is tuned in $\{5,10,20,\dots,100\}$ for the first two datasets and $\{5,10,20,\dots,50\}$ for WMT'14 De-En due to memory constraints. The best $\gamma$ is 0.9, 0.5, 0.5, for IWSLT'14 De-En, WMT'16 Ro-En, WMT'14 De-En, respectively; the best $b$ is 100, 70, 50 for the three datasets, respectively.

\paragraph{Details on $p_f$ and $p_r$.}

Recall that $p_f$ is the forward translator (from the source language to the target language) and $p_r$ is the reverse translator (from the target language to the source language). We use transformer-based architectures for all experiments. 
Refer to Table~\ref{tab:pf-details} for the architecture.

\paragraph{Number of parameters in the models.}
The IWSLT'14 De-En transformer has 39,469,056 parameters, the WMT'16 Ro-En transformer has 62,046,208 parameters, and the WMT'14 De-En transformer has 209,911,808 parameters.

\paragraph{Discussion on Q learning.}

In Section~\ref{sec:discussion}, to investigate whether better trajectories can improve Q learning results, we attempt adding high-reward trajectories from BSR as well as trajectories from a deep ensemble of two $p_f$'s. Deep ensembling two models (using different seeds) can produce high-quality translations. In this case, we simply want to use deep ensembling to diversify the sources of high-reward and high-BLEU trajectories. However, the result is that neither BLEU nor reward improves.

\section{Ethical Considerations}

IWSLT and WMT datasets are standard machine translation benchmarks. The datasets come from a variety of sources: phone conversations, parliament proceedings, news, and so on. There may be naturally occurring social biases in the datasets which have not undergone thorough cleansing. Training on these potential biases may lead to biased generations. There has been recent work studying such biases \citep{kocmi-etal-2020-gender}.

This work deals with speeding up inference, but not pretraining or training \citep{liu2021faster,hou2022token}. 
The standard practice of creating the pseudo-corpus requires a significant amount of computation. This step is optional, but it gives a boost in performance. We argue that if the MT system is put into production, then the benefit from the efficient inference will outweigh the cost of generating the pseudo-corpus.  

\end{document}